%% file: samplepaper.tex
\documentclass[runningheads]{llncs}
\usepackage{booktabs}
\usepackage{amsmath}
\usepackage{hyperref}
\usepackage{xcolor}
\usepackage[inline]{enumitem}
\usepackage{amssymb}
\usepackage[T1]{fontenc}
\usepackage{graphicx}
\usepackage{color}

\def\methodName{SURGIVID}

\begin{document}
\title{SURGIVID: Annotation-Efficient Surgical Video Object Discovery}

%
%
\author{Caghan Koksal \inst{1, 2} \and
Ghazal Ghazaei\inst{1} \and Nassir Navab  \inst{2,3}}
\authorrunning{C. Koksal, G. Ghazaei, N. Navab}
\institute{Carl Zeiss AG, Germany \and
Computer Aided Medical Procedures, Technical University of Munich, Germany \and 
Whiting School of Engineering, Johns Hopkins University, United States}
%
\maketitle 
\begin{abstract}
Surgical scenes convey crucial information about the quality of surgery. Pixel-wise localization of tools and anatomical structures is the first task towards deeper surgical analysis for microscopic or endoscopic surgical views. This is typically done via fully-supervised methods which are annotation greedy and in several cases, demanding medical expertise. 
Considering the profusion of surgical videos obtained through standardized surgical workflows, we propose an annotation-efficient framework for the semantic segmentation of surgical scenes.
We employ image-based self-supervised object discovery to identify the most salient tools and anatomical structures in surgical videos. These proposals are further refined within a minimally supervised fine-tuning step. Our unsupervised setup reinforced with only 36 annotation labels indicates comparable localization performance with fully-supervised segmentation models. Further, leveraging surgical phase labels as weak labels can better guide model attention towards surgical tools, leading to $\sim 2\%$ improvement in tool localization.  
Extensive ablation studies on the CaDIS dataset validate the effectiveness of our proposed solution in discovering relevant surgical objects with minimal or no supervision.

\end{abstract}

\begin{keywords}
Semantic Segmentation, Object Discovery, Minimal Supervision, 
\end{keywords}
\input{content/intro.tex}

\input{content/methods.tex}
\input{content/evaluation.tex}

\input{content/conclusion.tex}

\bibliographystyle{splncs04}
\bibliography{ref}

\end{document}

%% file: content/intro.tex
\section{Introduction}

Surgical videos encode a rich amount of information about surgical scenes, understanding of which can provide deeper insights into the quality of the surgery~\cite{al2019cataracts} as well as surgical workflow~\cite{blum2010modeling}. A spatio-temporal semantic understanding of the surgical scene can be leveraged for computer-assisted interventions~\cite{vercauteren2019cai4cai,nwoye2019weakly} as well as offline surgery analysis for surgical skill improvement~\cite{wang2018satr,hira2022video}, patient care optimization~\cite{mascagni2022artificial} and education. Considering the benefits and availability of abundant surgical videos for microscopic and endoscopic surgeries~\cite{lalys2014surgical,valderrama2022towards,zisimopoulos2018deepphase}, surgical procedures are still far from vast exploitation of surgical videos into their workflows. Besides the computational burden of video processing, video annotation is a major hindering factor for bringing surgical scene understanding solutions into practice. While pixel-wise annotation of surgical videos could be time-consuming and burdensome, requirement of medical expertise and additional temporal dimension can lead to increasing cost and difficulty of annotation. Although investigated broadly by both computer vision~\cite{cheng2022masked,chen2018encoder,ronneberger2015u} and medical~\cite{grammatikopoulou2019cadis,pissas2021effective,yuan2020object,chen2017deeplab} communities, the task of semantic segmentation is still among the most challenging tasks of image and video understanding. More recent advancements endeavor to tackle this problem by leveraging features of self-supervised foundation models such as DINO \cite{caron2021emerging}
LOST~\cite{simeoni2021localizing} and Tokencut~\cite{wang2022tokencut} leveraged DINO features towards self-supervised detection and segmentation models respectively. In LOST, authors proposed to use key feature vector of the last attention layer of DINO and calculate the patch similarities. Tokencut~\cite{wang2022tokencut} formulated object discovery as graph partitioning problem. The efficacy of this method is confined to predicting only the most salient object, thereby being incapable of predicting multiple objects. Mostly, some annotated labels are still needed to ensure a reliable semantic segmentation especially in the case of medical domain, which can benefit less from foundation models. Taking advantage of the foundation models~\cite{caron2021emerging}, recent developments in object discovery indicate huge progress in unsupervised object localization~\cite{wang2022tokencut,wang2023cut}. In this work, we introduce a minimal-annotation workflow for pixel-wise semantic segmentation in cataract surgery which can promote new possibilities for conducting annotations in the medical domain. Our contributions can be summarized as follows:

\begin{itemize}
    \item We successfully leverage the state-of-the-art in object discovery to localize the most influential components of the cataract surgery scene.
    \item We further exploit a self-training step using a Mask2Former model. This step refines unsupervisedly generated masks and identifies new masks corresponding to missing objects. Injecting minimal annotations into these identified masks leads to comparable pixel-wise localization accuracy to a fully-supervised method.
    \item We demonstrate that leveraging weak labels can effectively guide the model's attention towards the objects of interest within the scene.
    \item We highlight the efficacy of our annotation-efficient framework through comprehensive ablation studies on pixel-wise semantic segmentation of cataract videos, thereby emphasizing its impact.

\end{itemize}



%% file: content/methods.tex
\section{Methodology}
\subsection{Object Discovery}

Our backbone model features DINO~\cite{caron2021emerging}, a self-supervised model leveraging self-distillation, contrastive learning and Vision Transformers (ViT) ~\cite{dosovitskiy2020image}. DINO attention maps tend to be highly accurate indicating impressive potential in learning both local and global scene features which can be further exploited in downstream tasks such as semantic segmentation. 

In this work, we leveraged CutLER~\cite{wang2023cut} workflow to optimize the amount of annotations for semantic segmentation of surgical scenes within cataract surgery. Taking advantage of robust and informative attention of DINO, we generate class-agnostic pseudo segmentation masks using MaskCut~\cite{wang2023cut}. Each image is partitioned into equally-sized patches and fed into a pre-trained self-supervised DINO model to extract rich patch features represented by key vectors of the last attention layer. The cosine similarity between patches is then calculated to create an affinity matrix. \begin{equation} \label{eq:affinity_matrix}
    W_{i,j} = \frac{K_{i}K_j}{{\vert K_i \vert}_2{\vert K_j \vert}_2}
\end{equation}
where  \( K_i\)  and \( K_j\) are the key features of patch i and j, respectively.

Segmentation mask generation can be formulated as a graph partitioning task. Nodes of the graphs are the image patches and affinity matrix is used as an adjacency matrix. Then, the graph bi-partition is created by solving the N-cut problem. 
Bi-partitions are then converted to a foreground binary mask by selecting the bipartition that includes the most salient patch. After finding the first foreground object, the affinity matrix is re-masked to solve the N-cut problem on the masked affinity matrix. The above steps are iteratively followed until the number of maximum expected instances are reached.
As a second step, the generated unsupervised masks are used as pseudo-masks to train instance segmentation models.


\begin{figure}
    \centering
    \includegraphics[width=\textwidth,height=0.25\textwidth]{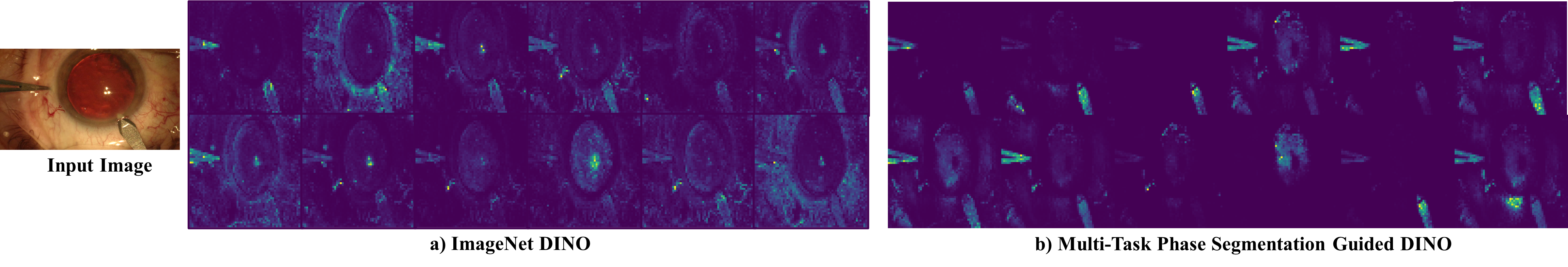}
    \caption{An illustration of DINO~\cite{caron2021emerging} attention maps for 12 different heads. a) Pretrained with ImageNet: Main anatomical structure and tools can be seen in the attention maps. b) Pretrained with ImageNet and fine-tuned on phase labels: attention maps are more focused on the surgical tools since tools include strong cues to detect the surgical phase of the surgery.}
    \label{fig:intro_figure}
\end{figure}

\begin{figure}
    \centering
    \includegraphics[width=\textwidth]{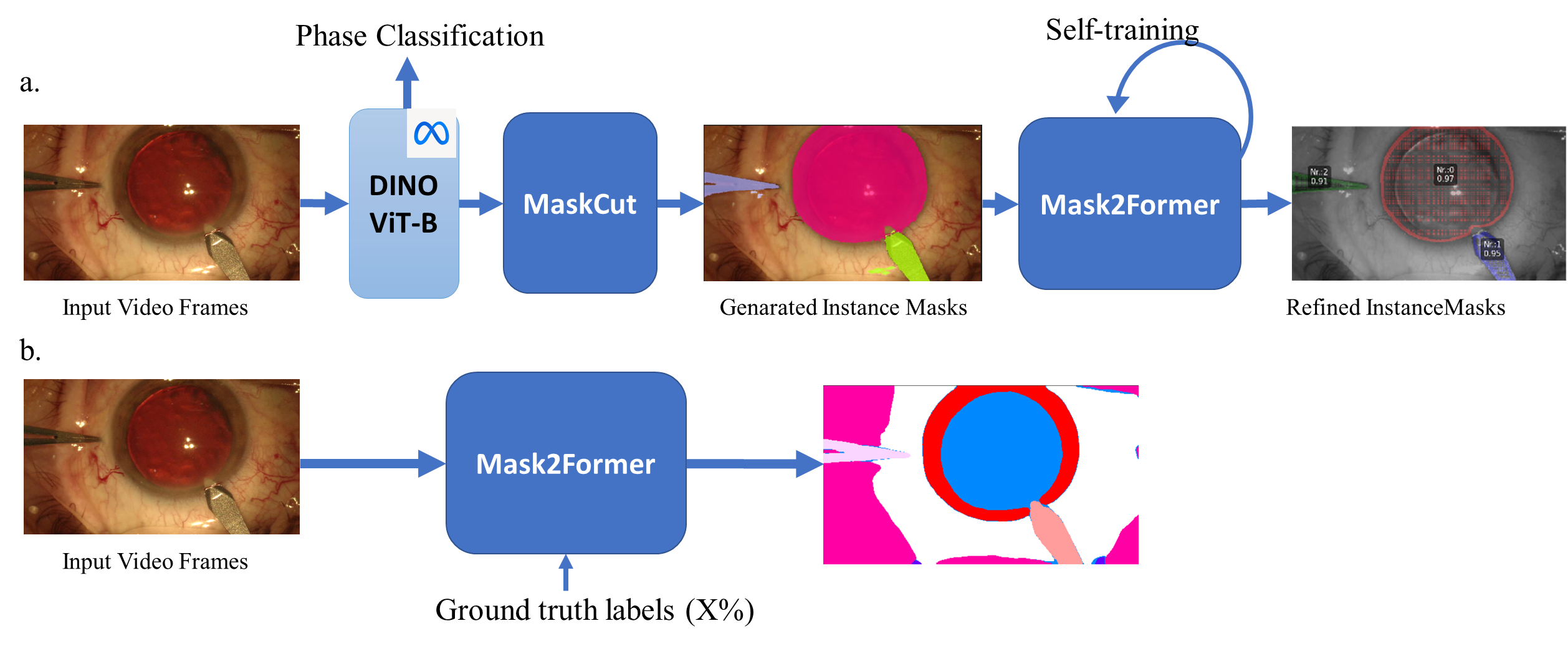}
    \caption{Overview of suggested workflow for unsupervised surgical scene segmentation: a. A pretrained DINO (optionally fine-tuned on phase labels in a multi-task learning setting) is used to extract rich scene features. Features are then leveraged within MaskCut~\cite{wang2023cut} to generate initial course masks of salient objects within the scene. The produced masks are further refined within a self-training step using Mask2former~\cite{cheng2022masked}. b. The Mask2Former pre-trained with pseudo-labels can then be fine-tuned with $X\%$ annotated masks.}
    \label{fig:method_figure}
\end{figure}
\subsection{DINO Weak Supervision}
To investigate the possibility of manipulating and further guiding the generated masks from MaskCut, we examined the usage of weak supervision from phase labels of the cataract surgery videos. We hypothesized that phase labels can deviate model attention to specific objects within the scene crucial for deciding on a phase, e.g. surgical tools. As such, we fine-tuned DINO in a multi-task learning scheme with surgical phase labels of the CATARACTS dataset~\cite{al2019cataracts}. We used $[CLS]$ token's feature and added linear layer on top to finetune DINO. Attention maps of surgical phase-guided DINO are illustrated in Figure ~\ref{fig:intro_figure} highlighting surgical instruments more elaborately. 
The training loss consists of surgical phase classification loss and DINO loss: $\mathcal{L}_{multi\_task}=\lambda_{cls}\mathcal{L}_{cls} + \lambda_{dino}\mathcal{L}_{dino}$ where $\mathcal{L}_{cls}$ and $\mathcal{L}_{dino}$ correspond to Cross Entropy and DINO  losses ~\cite{caron2021emerging} respectively and coefficients are selected as 0.5.
Finetuned DINO is used to generate features for MaskCut.
\subsection{Self-Training}
Mask2former offers a universal image segmentation framework which takes advantage of masked attention(Equation~\ref{eq:masked_attention}) enforcing local attention within the transformer decoder. This leads to faster convergence and better performance compared with cross-attention. 
\begin{equation} 
\label{eq:masked_attention}
    X_l = \text{softmax}(\mathbb{M}_{l-1} + Q_lK_l^T)V_l + X_{(l-1)}
\end{equation}
where ${X_l}\in\textbf{R}^{N \times C}$ corresponds to $N$ query features of layer $l$ of dimension $C$ and $Q_l =f_Q(X_{l-1})\in\mathbb{R}^{N\times C}$. $K_l$ and $V_l\in\mathbb{R}^{H_lW_l\times C}$  are key and value features for the image feature of resolution $H_l\times W_l$. $\mathbb{M}_{l-1} \in \{0,1\}^{N\times H_lW_l}$ is the attention mask.
In addition, Mask2former benefits from high-resolution features provided by its pixel decoder, which receives feature maps from a backbone ConvNet model and at a time feeds one scale of the features to the transformer decoder. Moreover, Mask2former query features are learned via a learnable positional embedding from zero initialization.
We use the realization and tiny \emph{Swin}~\cite{liu2021swin} backbone provided by \emph{hugging face}~\footnote{\url{https://huggingface.co/docs/transformers/model_doc/mask2former}}.

The training loss comprises of a mask loss as well as class loss $\mathcal{L}=\mathcal{L}_{mask}+\lambda_{cls}\mathcal{L}_{cls}$ where $\mathcal{L}_{cls}=\lambda_{ce}\mathcal{L}_{ce}+\lambda_{dice}\mathcal{L}_{dice}$. $\mathcal{L}_{ce}$ and $\mathcal{L}_{dice}$ corresponding to Cross Entropy and dice loss~\cite{milletari2016v} respectively.

We leveraged the MaskCut~\cite{wang2023cut} method and generated pseudo segmentation masks on the training split of the CaDIS dataset\cite{grammatikopoulou2019cadis}. DINO-B8 \cite{caron2021emerging} is used to extract features from the scenes of cataract surgeries. The pseudo masks from MaskCut~\cite{wang2023cut} are then exploited to train Mask2Former in a class-agnostic setting as a pre-training step. We named this pretraining step as self-training and further training on a labeled subset of the CaDIS as fine-tuning steps. In self-training, we trained Mask2Former for 10 epochs and chose the model with the lowest validation error. Then, we fine-tuned the self-trained Mask2Former on various amounts of annotation labels from the CaDIS dataset.

%% file: content/evaluation.tex
\section{Evaluation}

\subsection{Dataset}
CaDIS~\cite{grammatikopoulou2019cadis} is a subset of CATARACTS~\cite{al2019cataracts} video dataset, which comprises a collection of sampled frames from various videos. In total, CaDIS contains 4,670 images that have been annotated pixel by pixel. These annotations are based on 36 semantic classes, which encompass 29 different surgical instruments, 4 anatomy classes, and 3 classes for other external objects. CaDIS defines three semantic segmentation tasks, each with increasing levels of detail. For our purposes, we focus on TaskII, which strikes a balance in granularity and offers labels for each class. TaskII consists of 17 classes, with 9 of them corresponding to surgical instruments.
 During fine-tuning stage, only training set is subsampled to 1\%, 10\%, 50\% and they subsets contain 36, 355, 1775 samples respectively. Original val and test splits are kept for each experiments.


\subsection{Metrics}
We evaluate the semantic segmentation models based on the mean intersection over union (mIoU) and pixel-wise accuracy enabling both spatial and semantic evaluation of the results. The localization performance of unsupervised models is calculated by matching predicted masks with ground truth masks using Intersection over Union (IoU). The best match is calculated by optimizing the Hungarian algorithm. To measure the mIoU$_{loc}$ the ground truth class labels are ignored to be able to assess only the localization performance. This metric is specifically more interesting in the unsupervised stage, where instances do not belong to any class IDs.

 

\subsection{Implementation Details}


\subsubsection{Self-training and Weak Supervision}
In the self-training phase, we used Mask2Former with Swin-tiny and trained it on the pseudo-masks generated by MaskCut.After hyperparameter search, we decided to use 5 pseudo masks with MaskCut per image. Mask2Former is trained only for 10 epochs to not overfit on pseudo masks considering the noisy MaskCut output. The batch size was set to 16 and horizontal flipping augmentation was used during self-training. DINO is finetuned using the  default hyperparameters.
    
\subsubsection{Fine-tuning}
After pre-training, Mask2Former is fine-tuned on the CaDIS training set. To evaluate the annotation efficiency of the proposed solution, we performed data ablation experiments on a limited subset of annotated data. Our experiments include 4 subsets of the dataset progressively increasing the annotation amount. We randomly sample 1\%, 10\%, 50\% of the CaDIS training set. In  the 10\%, 50\%, 100\% cases, the model is trained for 50 epochs, and best model is selected on the validation set. In the 1\% case, the model is trained for 200 epochs. AdamW\cite{loshchilov2017decoupled} with an initial learning rate 1e-4 is used in all experiments. We used a single RTX 3090 with 24 GB memory for all experiments.


\subsection{Results and Discussion}

\subsubsection{Mask2Former}
The Mask2Former with a Swin backbone was trained on the training set of CaDIS dataset to provide a baseline for the results. As results suggest, Mask2Former performs slightly better than OCRNet~\cite{pissas2021effective} which is the state of the art on CaDIS dataset. OCRNet creates a soft segmentation map by dividing the contextual pixels into a set of soft object regions each assigned to a class. We continued experiments using Mask2Former thanks to its prior performance in tool segmentation, which could be more challenging to localize and convey more crucial information for surgical scene understanding.

\begin{table}[h!]
    \centering
    \caption{Results for pixel-wise semantic segmentation on Task~II of CaDIS~\cite{grammatikopoulou2019cadis} test set}
     \resizebox{\textwidth}{!}{%
    \begin{tabular}{lc|c|c|c}
        & Res & mIoU(Anatomy) & mIoU (Instr.) & mIoU (All Classes) \\
        \toprule
        HRNetV2~\cite{grammatikopoulou2019cadis} & 270x480 & - & -&76.1 \\
        OCRNet~\cite{pissas2021effective} & 540x960 & 85.92 & \textbf{73.49} &79.09 \\
        \midrule
        Mask2Former~\cite{cheng2022masked} & 270x480 & \textbf{86.36} & 65.91 & 79.91 \\
        Mask2Former~\cite{cheng2022masked} & 540x960 & 86.07 & 68.53 & 80.51 \\
        Ours & 270x480  & \textbf{86.36} & 69.55 & \textbf{80.69} \\
        \bottomrule
    \end{tabular}
    }
    \label{tab:supervised_comparison}
\end{table}
Comparison of our method to SOTA is summarized in Table~\ref{tab:supervised_comparison}. Supervised Mask2Former overperforms the SOTA method. Moreover, our workflow improves the mIOU and achieves the SOTA mIOU result on CaDIS Task 2.
\subsubsection{MaskCut}
 Figure~\ref{fig:unsupervised} manifests initial course masks predicted by MaskCut for various scenes with different levels of complexity, e.g. multiple tools (Figure~\ref{fig:unsupervised}.a,d), transparent (Figure~\ref{fig:unsupervised}.c) and narrow tools (Figure~\ref{fig:unsupervised}.b). It can be seen that in most cases, MaskCut discovers at least one prominent object with correct localization. A difficulty for MaskCut seems to be narrow tools and differentiation among multiple instances especially when there is no clear distinction between the boundaries of components.


\subsubsection{Self-training with Mask2Former}
\begin{figure} [h!]
    \centering
    \includegraphics[width=\textwidth]{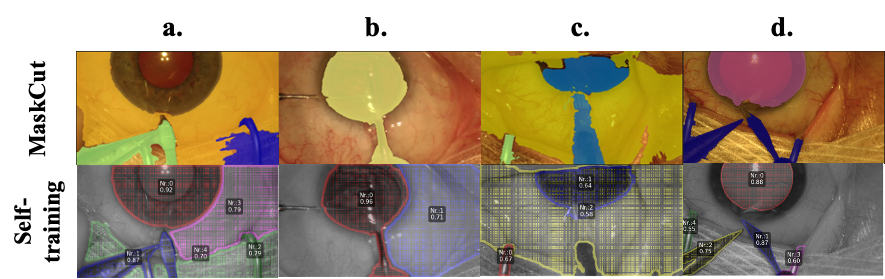}
    \caption{Qualitative results of the unsupervised steps of suggested framework, namely the MaskCut and self-training. Self-training increases the segmentation quality and helps to find tools that are not detected with MaskCut. }
    \label{fig:unsupervised}
\end{figure}

We compare the masks generated by self-training to validate whether the course masks are further refined during training with pseudo-masks. As Figure~\ref{fig:unsupervised} illustrates, the challenging instances for MaskCut are now differentiated (Figure~\ref{fig:unsupervised}.d). Although in Figure~\ref{fig:unsupervised}.a the tool instances are not refined, the Cornea instance is added. Also in the case of Figure~\ref{fig:unsupervised}.c it can be observed that the MaskCut discovers the transparent lens injector while the mask refinement reduces the instance mask and separates the Cornea. 
\subsubsection{Fine-tuning with Mask2Former}
Finally, an ablation study was designed to study the impact of annotation on the performance of a self-trained model, the results of which are summarized in Table~\ref{tab:ablations}. Our results highlight the effectiveness of the proposed unsupervised step especially for the model with 1\% annotation only indicating an accuracy and mIoU$_{loc}$ comparable with that of a fully-supervised model with abundant samples of annotation. The 1\% annotation model falls short in the mIoU score, stemming from the lack of exposure to a balanced set of semantic classes. That is, the model may miss primary tools such as primary knife, lens injector, etc. as it may not encountered these tools at all during training. However, 1\% of data is sufficient to segment anatomical structures as the anatomy classes barely vary along the surgery. As the quantity of training samples increases, the boundaries become smoother, noise is reduced, and the efficacy of class prediction improves.
Finally, as the Table~\ref{tab:ablations} suggests, injection of weak supervision via phase labels during DINO fine-tuning indicates better overall IoU and more specifically $	\sim 2\%$ improvement of tool localization. That is, by incorporating phase segmentation as an additional supervisory task in the multi-task DINO, our proposed method focuses model attention towards surgical instruments. This is due to the fact that the type and movement of surgical tools contribute to improved differentiation of surgical phases. These observations can be used further for guided localization of objects of interest using weak supervision.


\begin{figure} [h]
    \centering
    \includegraphics[width=\textwidth]{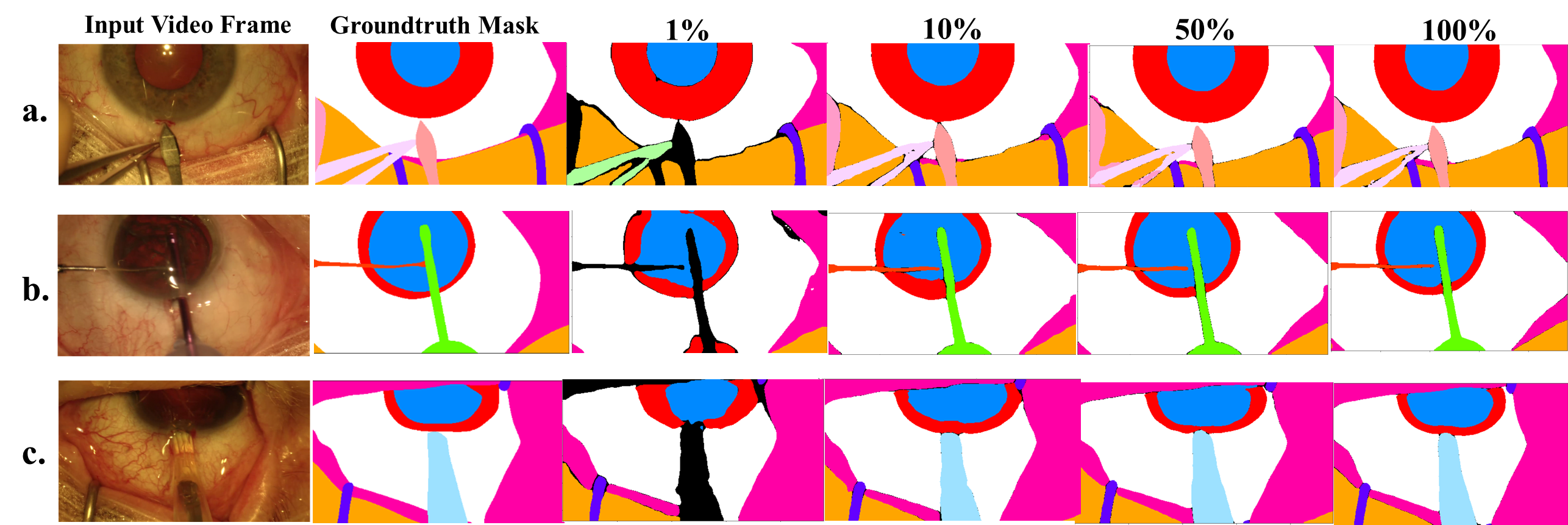}
    \caption{Qualitative results on Task~II indicating the impact of annotation via feeding gradually increasing portions of labels to the self-trained model. }
    \label{fig:qualitative}
\end{figure}

\begin{table} [h!]
    \centering
    
    \caption{Semantic segmentation results of used models on TaskII of CaDIS~\cite{grammatikopoulou2019cadis} test set. *: refers to surgical phase guided DINO model.}
     \resizebox{\textwidth}{!}{%
    \begin{tabular}{c c  | |c c c c c c c c c} 
        \toprule
         \multicolumn{2}{c}{Supervision} & \multicolumn{3}{c}{Model} &  \multicolumn{6}{c}{Metrics} \\
         \toprule
         DINO & Subset & Maskcut & M2F & PAC & mIoU &  mIoU$_{ana}$ &mIoU$_{ins}$  & mIoU$_{loc}$ &AP@50 & AP@75\\
         
         ImageNet & 0\%    & \checkmark & \checkmark  & - & - & - & -  & 21.35  & - & - \\
         \midrule
         ImageNet  & 1\%  & \checkmark & \checkmark   &92.73 &71.83  &84.12 & 43.33 & 76.29 & 56.90 &38.65\\
         ImageNet  & 10\%  &  \checkmark & \checkmark  &93.90  &75.88 & 85.80 & 52.32& 80.69 & 72.09 & 53.53 \\
         ImageNet  & 50\%  & \checkmark & \checkmark  &94.57  &80.32 &\textbf{86.52} &67.36 & 82.54 & 81.62 &  61.45 \\
         ImageNet  & 100\%  & \checkmark & \checkmark &\textbf{94.62}&80.54 &86.11 &67.80 & 82.58 & 80.61 & 62.50 \\
         ImageNet+Phase *  & 100\%  & \checkmark & \checkmark &93.85 &\textbf{80.69} & 86.36 &\textbf{69.55} & \textbf{82.82} & \textbf{83.11} & \textbf{63.44} \\
         \midrule
         M2F Baseline  & 100\%  & - & \checkmark &94.51 &79.91 &86.36 &65.91 & 81.56 & 81.37 & 59.96 \\

         \bottomrule
    \end{tabular}
    }
\label{tab:ablations}
\end{table}

%% file: content/conclusion.tex
\newpage
\section{Conclusion}
In this work, we introduce \methodName{}. This annotation-efficient workflow tackles the annotation-burdensome task of semantic segmentation for surgical videos by 
bringing together the capabilities of the state-of-art self-supervised object discovery method, CutLER,  and segmentation model, Mask2Former. Through extensive evaluations, we demonstrate that training a class-agnostic segmentation model on generated pseudo-masks improves the quality of the masks with no labels. By reducing the labels up to 90\%, we demonstrate on-par performance with a fully-supervised setup on primary tools and anatomies within a surgical scene of CaDIS dataset. Furthermore, by leveraging weak supervision from cataract surgery phase labels, more accurate localization of instruments can be achieved.
Our proposed workflow has the potential to facilitate the utilization of large volumes of unlabeled surgical videos by the medical community. As future work, it would be valuable to explore the application of our method to a broader range of surgical procedures, in order to assess its generalizability and effectiveness in diverse clinical contexts.